\documentclass[11pt]{article} 
\usepackage[english]{babel}
\usepackage[utf8]{inputenc}
\usepackage[T1]{fontenc}
\usepackage{latexsym}
\usepackage{url}
\usepackage[dvips]{graphicx}
\usepackage[normalem]{ulem}
\usepackage{amsmath,amsfonts,amssymb,amsthm,bm}
\usepackage{psfrag}
\usepackage{fullpage}
\usepackage{subcaption}
\usepackage{enumitem}
\usepackage{color}
\usepackage[table]{xcolor}
\usepackage[font={small}]{caption}
\usepackage[output-decimal-marker={,},exponent-product=\cdot]{siunitx}
\usepackage[hang,flushmargin]{footmisc}
\usepackage{hyperref}

\usepackage{booktabs}
\usepackage[section]{placeins}
 \usepackage[ruled, vlined]{algorithm2e}
\usepackage{algorithmic}
\usepackage{microtype}
\usepackage{xspace}
\usepackage{complexity}
\usepackage{mathtools, nccmath}
\usepackage[misc]{ifsym}
\DeclareMathOperator*{\argmax}{arg\,max}

\newcommand{\noi}{\noindent}

\mathchardef\mhyphen="2D
\newtheorem{lemma}{Lemma}
\newtheorem{proposition}{Proposition}
\newtheorem{corollary}{Corollary}
\newtheorem{assumption}{Assumption}
\newtheorem{example}{Example}

\def\noi{\noindent}

\numberwithin{equation}{section}

\usepackage{mleftright} % for '\mleft' and '\mright' macros
\renewcommand{\qed}{\hfill\ensuremath{\blacksquare}}

\small \normalsize

\begin{document}

\title{Algorithms for slate bandits with non-separable reward functions}

\author{
{Jason Rhuggenaath (\Letter)}\thanks{
 Eindhoven University of Technology, Eindhoven, The Netherlands, e-mails: {\tt $\{$j.s.rhuggenaath, a.e.akcay, yqzhang, u.kaymak$\}$@tue.nl}}
\and {Alp Akcay}\footnotemark[1]
\and {Yingqian Zhang}\footnotemark[1]
\and {Uzay Kaymak}\footnotemark[1]
}

\date{This version: April 20, 2020}

\maketitle
\sloppy

\begin{abstract} 

\noi In this paper, we study a slate bandit problem where the function that determines the slate-level reward is non-separable: the optimal value of the function cannot be determined by learning
the optimal action for each slot. We are mainly concerned with cases where the number of slates is large
relative to the time horizon, so that trying each slate as a separate arm in a traditional multi-armed bandit, would not be feasible. Our main contribution is the design of algorithms that still have sub-linear regret with respect to the time horizon, despite the large number of slates.
Experimental results on simulated data and real-world data show that our proposed method outperforms popular
benchmark bandit algorithms.
\medskip\medskip

\noi \textbf{Keywords:} multi-armed bandits, slate bandits, combinatorial bandits.

\end{abstract} %\vspace*{1em}

\section{Introduction}
\label{sec:Introduction}
In many practical problems an agent needs to choose an action from a set where each action leads to a random reward.
The objective is to devise a policy that maximizes expected cumulative rewards over a finite time horizon. Often the reward distribution is unknown, and as a consequence, the agent faces an exploration-exploitation
trade-off. The multi-armed bandit problem ~\cite{lai1985asymptotically,MAL-024} is a standard framework for studying such exploration-exploitation problems.

Many problems in the domain of web-services, such as e-commerce, online advertising and streaming, require the agent to select not only one but multiple actions at the same time. After the agent makes a choice, a collective reward characterizing the quality of the entire selection is observed. Problems of this type are
typically referred to as slate bandits or combinatorial bandits ~\cite{CESABIANCHI20121404,NIPS2015_5831_short}. In a slate bandit problem, a slate consists of a number
of slots and each slot has a number of base actions. Given a particular action for each slot, a reward function
defined at the slate-level determines the reward for each slate.

% Slate bandits can arise in a range of applications.
One example of a slate bandit problem is the design of recommender systems for a streaming service such as Netflix. When a user logs in to his account, the streaming service displays a page that recommends various movies and shows. This can be interpreted as a slate bandit problem where the slots are the different genres (e.g. comedy, romance, etc.) and the base actions are the titles in each genre. The goal is to recommend a set of titles such that  the probability of a user playing something from the set is maximized. 

Another example is the reserve price optimization problem on a header bidding platform (see Section~\ref{RP_HB_application} for more details). In this problem, a publisher has to select a reserve price for each partner on the header bidding platform and the revenue at the selected reserve price is stochastic. The revenue for the publisher is the maximum of the revenues of all of the partners on the header bidding platform.

Previous studies (see e.g. \cite{Kale:2010:NBS:2997189.2997307,qin2014contextual,Dimakopoulou}) assume that the reward function at the slate level is additive or that the expected reward at the slate-level is a non-decreasing function of the expected rewards at the slot-level (this is also called the monotonicity assumption). This implies that the optimal action at the slate level can be found by finding the optimal base action for each individual slot. In some applications the monotonicity assumption might be reasonable, but in some cases it might not apply. For example, if the slate-level reward is the maximum (or minimum) of the rewards at the slot level, then the monotonicity assumption is no longer satisfied. This is an example of a non-separable slate-level reward function.
The reserve price optimization problem mentioned above is thus a concrete example of a slate bandit problem with a non-separable reward function. Another example arises in maintenance and reliability problems, where the failure time of a system is the minimum of a set of random variables. 

In this paper we study slate bandits with  non-separable reward functions.
To the best of our knowledge, this variant of the slate bandit problem has not been studied before and existing algorithms cannot be applied.  
We are mainly concerned with cases where the number of slates is large
relative to the time horizon, so that trying each slate as a separate arm in a traditional multi-armed bandit, would not be feasible. In such cases it is not immediately clear whether sub-linear regret is possible, and therefore we study the design of algorithms that have sub-linear regret.
We summarize the main contributions of this paper as follows:
\begin{itemize}
\item  To the best of our knowledge, we are the first to study slate bandits with non-separable reward functions.
\item We provide a theoretical analysis and derive problem-dependent and problem-independent regret bounds.  We provide   algorithms that  have sub-linear regret with respect to the time horizon. 
\item Experimental results on simulated data and using real-world data show that our proposed method outperforms popular benchmark bandit algorithms.
\end{itemize}

The remainder of this paper is organized as follows. In Section \ref{sec:Literature} we discuss the related literature. Section \ref{sec:problem_formulation} provides a formal formulation of the problem. In Section \ref{sec:algorithm} we present the our proposed algorithms for the slate bandit problem and provide a theoretical analysis. 
In Section~\ref{sec:experiments} we perform experiments and compare our method with baseline strategies in order to assess the quality of our proposed algorithms.  
Section~\ref{sec:conclusion} concludes our work and  provides some interesting directions for further research. 

\section{Related Literature}
\label{sec:Literature}
The  slate bandit problem  has been studied before in multiple prior papers and  these papers study  different variants of the   problem and make different assumptions. 
The main variants of the slate bandit problem center around three properties of the problem: (i) whether the  slot-level rewards in the slate are observed or not (the situation where the slot-level rewards are observed is often referred to as semi-bandit feedback in the literature); (ii) whether the function that  determines the slate-level reward is known or not; (iii) the structural properties of the function that determines the slate-level reward.

In \cite{Kale:2010:NBS:2997189.2997307,CESABIANCHI20121404,Li:2016:CCC:3045390.3045522,qin2014contextual,pmlr-v28-chen13a_short,Wen:2015:ELL:3045118.3045237,pmlr-v38-kveton15_short,NIPS2015_5831_short,Wang:2017:EOC:3298483.3298634}  slate bandits  with semi-bandit feedback are studied. In  \cite{Kale:2010:NBS:2997189.2997307,NIPS2017_6954_short,Wen:2015:ELL:3045118.3045237,pmlr-v38-kveton15_short} it is assumed that the slate-level reward is an additive function of the rewards of the individual slots. In \cite{NIPS2017_6954_short} the slot-level rewards are assumed to be  unobserved, while \cite{Kale:2010:NBS:2997189.2997307} assumes that slate-level reward function is known. Some papers make other structural assumptions about the slate-level reward function. In \cite{qin2014contextual,pmlr-v28-chen13a_short,Li:2016:CCC:3045390.3045522,Chen_slate_2016} two key structural assumptions are made: a monotonicity assumption and a bounded smoothness (or Lipschitz continuity) assumption. In addition, \cite{qin2014contextual,pmlr-v28-chen13a_short,Li:2016:CCC:3045390.3045522} do not assume that the slate-level reward function is known. Instead, they assume that an \( \alpha\)-approximation oracle is available.

In related work  \cite{Dimakopoulou} do not assume that the slot-level rewards are observed and  that the slate-level reward function is known. They exploit a monotonicity assumption (similar to \cite{qin2014contextual,pmlr-v28-chen13a_short,Li:2016:CCC:3045390.3045522,Chen_slate_2016}) that relates the  slot-level rewards to the slate-level rewards and propose an  approach based on Thompson sampling in order to balance exploration and exploitation. 

The main difference between our paper and the aforementioned works, is that we do not assume that the slate-level reward is additive or that the expected reward at the slate-level is a non-decreasing function of the expected rewards at the slot-level. However, unlike in \cite{Dimakopoulou},  we assume that the slate-level reward function is known. Furthermore, we do not make use of approximation oracles as in \cite{qin2014contextual,pmlr-v28-chen13a_short,Li:2016:CCC:3045390.3045522,Chen_slate_2016}. 

To the best of our knowledge, this variant of the slate bandit problem has not been studied before and existing algorithms cannot be applied.  
We are mainly concerned with cases where the number of slates is large
relative to the time horizon, so that trying each slate as a separate arm in a traditional multi-armed bandit, would not be feasible. In such cases it is not immediately clear whether sub-linear regret is possible. The main contribution of this paper is the design of algorithms that still have sub-linear regret with respect to the time horizon, despite the large number of slates.

\section{Problem formulation} \label{sec:problem_formulation}
  \subsection{Problem definition and notation} 

We consider a slate bandit problem that is similar to \cite{Dimakopoulou} and the unordered slate bandit problem in \cite{Kale:2010:NBS:2997189.2997307}.  The set of actions (the slates) is given by \( \mathcal{B}\) with  \( |\mathcal{B}| = \bar{K}  \).  If action \( b \in \mathcal{B} \) is selected, then the reward is a random variable \(  Y(b)  \).  
A slate consists of \( M \in \mathbb{N} \) slots, where \( M >1 \).  Each action is a vector in \( \mathbb{R}^{M} \). That is, \(  b \in \mathbb{R}^{M} \) for all \( b \in \mathcal{B} \).  Slot \( i \in \lbrace 1, \dots, M \rbrace\) has a set of base actions \( \mathcal{B}_{i} \) with \( |\mathcal{B}_{i}| = K_{i} \).  The set of slates \( \mathcal{B} \) is given by \( \mathcal{B} =  \mathcal{B}_{1} \times \mathcal{B}_{2} \times \dots \times \mathcal{B}_{M}  \). 
We make the following assumptions regarding the slot-level action sets.

\begin{assumption} \label{ass_ordered_prices_B_i}
Without loss of generality we assume  that \( \mathcal{B}_{i} = \lbrace 1, \dots, K_{i}\rbrace \)  for \( i = 1, \dots, M \).
\end{assumption}

\begin{assumption} \label{ass_num_elements_B_i}
Without loss of generality we assume that  \( |\mathcal{B}_{i}| = K_{i} = K  \)   for \( i = 1, \dots, M \).
\end{assumption}

 Given an action \( b \in \mathcal{B} \) the random variable \(  Y(b)  \) satisfies \(  Y(b)   = f(Y_{1}(b_{1}), \dots,  Y_{M}(b_{M})) \), where \( b_{i} \in \mathbb{R} \) is the \(i\)-th element of  action \( b  \) and where \( Y_{i}(b_{i}) \) for \( i = 1, \dots, M \) is a random variable.  We make the following assumptions regarding the slate-level reward function.

\begin{assumption} \label{ass_independent_components_B}
Let \( b \in \mathcal{B} \) and \(  Y(b)   = f(Y_{1}(b_{1}), \dots,  Y_{M}(b_{M})) \). Then, \( Y_{i}(b_{i}) \) is independent from \( Y_{j}(b_{j}) \) for all \( j \neq i \). 
\end{assumption}

\begin{assumption} \label{ass_bounded_rewards}
The rewards are bounded and such that  \( Y_{i}(\cdot) \in [0,1] \) for all \( i \in \lbrace 1, \dots, M \rbrace \).  
\end{assumption}

\begin{assumption} \label{ass_mapping_f}
The function \( f \)   is known and satisfies \(  f: \mathbb{R}^{M} \rightarrow [0,1]  \).
\end{assumption}

For \( b \in \mathcal{B}  \)  define the quantity \( \mu{(b)} =  \mathbb{E} \left \lbrace  Y(b) \right\rbrace\) and let \( b^{*} = \argmax_{b \in \mathcal{B}} \mu{(b)} \). 
The optimality gap for action \(b \in \mathcal{B}  \) is defined as \(   \Delta(b) =\mu{(b^{*})} -  \mu{(b)} \).   
Define \( \Delta_{\min} = \min \lbrace  \Delta(b) | b \in \mathcal{B}, b \neq b^{*} \rbrace \). Here \( \Delta_{\min} \) measures the optimality gap between the best action and the second-best action. We assume that the optimality gaps satisfy \( \Delta_{\min} \geq \varepsilon > 0 \) for some  \( \varepsilon \in \mathbb{R} \). This assumption enforces that the optimality gap is bounded from below and ensures that the notion of `the best action' and `the second-best action'  is well-defined.

 We assume that the decisions are implemented according to the following online protocol: for each round \( t \in \lbrace 1, \dots, T \rbrace\)
\begin{enumerate}
    \item the agent selects a slate \( b \in \mathcal{B} \).
     \item the agent observes \( Y_{i}(b_{i}) \) for \( i = 1, \dots, M \).  The agent receives \(r_{t} \sim  Y(b)  \) where \(  Y(b)   = f(Y_{1}(b_{1}), \dots,  Y_{M}(b_{M})) \). The rewards \( r_{t} \) are independent over the rounds.  
    %  and \(Y(b_{1}) \) is independent of \(Y(b_{2}) \)   for \( b_{1} \neq b_{2}  \).
\end{enumerate}

\noindent For a fixed sequence \( i_{1}, \dots, i_{T} \) of selected actions, the pseudo-regret over \( T \) rounds is defined as \( R_{T} = T \cdot \mu^{*} - \sum_{t = 1}^{T} \mu{(i_{t})} \).
 The expected pseudo-regret is defined as \(\mathcal{R}_{T} =  \mathbb{E} \left \lbrace R_{T}  \right\rbrace \), where the expectation is taken with respect to possible randomization in the selection of the actions \( i_{1}, \dots, i_{T} \). 
 
The slate bandit problem is challenging due to the number of actions growing exponentially in \(M \), and due to the non-separable reward function which implies that a local optimization of
slate rewards does not necessarily imply a global optimum. That is,  \( \mathbb{E} \left \lbrace Y(b) \right\rbrace \) cannot  necessarily be maximized by choosing the action with the  highest expected reward at the slot-level for each slot.
Note that we allow for an arbitrary function \( f \) in Assumption~\ref{ass_mapping_f} and that the reward distributions at the slot-level can also be arbitrary (as long as they are bounded in \( [0,1] \)). Existing papers assume that \( f \) is  an additive function  or that \( \mathbb{E} \left \lbrace Y(b) \right\rbrace \)  satisfies a monotonicity property. In Example~\ref{example_monotonic} below we give a concrete example that shows that existing algorithms that exploit this monotonicity property (such as \cite{Dimakopoulou,qin2014contextual,pmlr-v28-chen13a_short,Li:2016:CCC:3045390.3045522,Chen_slate_2016}) can fail to learn the best slate. Therefore, existing algorithms are in general not guaranteed to solve our problem. Assumption~\ref{ass_independent_components_B} may seem restrictive, but even under this assumption, this problem is still non-trivial and, to the best of our knowledge, there are no other existing algorithms to solve this problem. Note that Example~\ref{example_monotonic} shows that, even under Assumption~\ref{ass_independent_components_B}, existing algorithms can fail to learn the best slate.

\begin{example} \label{example_monotonic}
Consider a simple instance of the slate bandit problem where there are \( M = 2 \) slots. Let \( \mathcal{B}_{1} = \lbrace a , b  \rbrace  \), \( \mathcal{B}_{2} = \lbrace c , d  \rbrace  \). Let \( Y_{1}(a) \sim U(0.4,0.5) \), \( Y_{1}(b) \sim U(0.0,0.1) \),   \( Y_{2}(c) \sim U(0.4,0.5) \), \( Y_{2}(d) \sim U(0.15,0.7) \). Here \( U(v,w) \) denotes a uniform distribution on \( [v, w] \).
% Assume that  \( Y_{1}(a)\), \( Y_{1}(b) \), \( Y_{2}(c)\), \( Y_{2}(d) \) are all independent of each other.  
For each slot, there are 2 actions. There are 4 slates in total and the slates are given by \( \mathcal{B} = \lbrace \lbrace a,c \rbrace , \lbrace a,d \rbrace , \lbrace b,c \rbrace , \lbrace b,d \rbrace   \rbrace \). \\
\noindent The  rewards at the slate level are given by:
\begin{align*}
Y(\lbrace a,c \rbrace) = \max \lbrace Y_{1}(a), Y_{2}(c) \rbrace    \\
Y(\lbrace a,d \rbrace) = \max \lbrace Y_{1}(a), Y_{2}(d) \rbrace  \\
Y(\lbrace b,c \rbrace) = \max \lbrace Y_{1}(b), Y_{2}(c) \rbrace  \\
Y(\lbrace b,d \rbrace) = \max \lbrace Y_{1}(b), Y_{2}(d) \rbrace 
\end{align*} 

% \noindent Now it is not difficult to show that 
% \( \mathbb{E} \left \lbrace Y(\lbrace a,d \rbrace) \right\rbrace  >  \mathbb{E} \left \lbrace Y(\lbrace a,c \rbrace) \right\rbrace \). \\
\noindent Let \( \mu_{a} =  \mathbb{E} \left \lbrace Y_{1}(a) \right\rbrace\), \( \mu_{b} =  \mathbb{E} \left \lbrace Y_{1}(b) \right\rbrace\), \( \mu_{c} =  \mathbb{E} \left \lbrace Y_{2}(c) \right\rbrace\) and \( \mu_{d} =  \mathbb{E} \left \lbrace Y_{2}(d) \right\rbrace\). 

\noindent Existing algorithms in \cite{Dimakopoulou,qin2014contextual,pmlr-v28-chen13a_short,Li:2016:CCC:3045390.3045522,Chen_slate_2016} make a monotonicity assumption. This assumption states that if the vector of mean rewards of the slots in  a slate (say slate \(A \)) dominates the vector of mean rewards of the slots in another slate (say slate \(B \)), then the expected reward of slate \( A \) is at least as high as the expected reward of slate \(B \). In this example the monotonicity assumption implies that, if  \( \mu_{c} \geq \mu_{d} \), then it must be that \( \mathbb{E} \left \lbrace Y(\lbrace a,d \rbrace) \right\rbrace  \leq  \mathbb{E} \left \lbrace Y(\lbrace a,c \rbrace) \right\rbrace \). 

\noindent  Note that from the properties of the uniform distribution we have that \( \mathbb{E} \left \lbrace Y_{1}(a) \right\rbrace =  \mathbb{E} \left \lbrace Y_{2}(c) \right\rbrace > \mathbb{E} \left \lbrace Y_{2}(d) \right\rbrace\).
Note that the monotonicity assumption that is used in  \cite{Dimakopoulou,qin2014contextual,pmlr-v28-chen13a_short,Li:2016:CCC:3045390.3045522,Chen_slate_2016} implies that we should have \( \mathbb{E} \left \lbrace Y(\lbrace a,d \rbrace) \right\rbrace  \leq  \mathbb{E} \left \lbrace Y(\lbrace a,c \rbrace) \right\rbrace \). However, it can be shown that \( \mathbb{E} \left \lbrace Y(\lbrace a,d \rbrace) \right\rbrace  >  \mathbb{E} \left \lbrace Y(\lbrace a,c \rbrace) \right\rbrace \) in this example. Therefore,  the monotonicity assumption  implies that slate \( \lbrace a,c \rbrace \) has an expected reward that is at least as high as the expected reward of slate \(\lbrace a,d \rbrace \) and this implication is false. Existing algortihms that rely on the  monotonicity assumption are therefore not guaranteed to learn the best action in this slate bandit problem.  \qed
\end{example}

\subsection{Example application: reserve price optimization and header bidding} \label{RP_HB_application}

One of the main mechanisms that web publishers use in online advertising in order to sell their advertisement space  is the real-time bidding (RTB) mechanism \cite{RTB_survey}. In RTB there are three main platforms: supply side platforms (SSPs), demand side platforms (DSPs) and an ad exchange (ADX) which connects SSPs and DSPs. The SSPs collect inventory of different publishers and thus serve the supply side of the market. Advertisers which are interested in showing online advertisements are connected to DSPs.  A real-time auction  decides which advertiser is allowed to display its ad and the amount that the advertiser needs to pay. Most of the ad inventory is sold via second-price auctions with a reserve price \cite{RTB_survey,Mohri:2016:LAS}. In this auction, the publisher  specifies a value \( p_{t} \) (the reserve price) which represent the minimum price that he wants for the impression. The revenue for the publisher (at a particular reserve price) is random and depends on the highest bid (\(X_{t}\)) and second highest bid (\(W_{t}\)) in the auction. The revenue   of the publisher in round \(t \) is given by \( R_{t}(p_{t}) = \mathbb{I}\lbrace  p_{t} \leq X_{t} \rbrace \cdot \max\lbrace W_{t}, p_{t}  \rbrace \). 

In header bidding (see e.g. \cite{Jauvion:2018_TS}), the publisher can connect to multiple SSPs for a single impression. The publisher specifies a reserve price for each SSP and each SSP runs a second-price auction. After the SSPs run their auctions, they return a value back to the header bidding platform indicating the revenue for the publisher if they sell on that particular SSP. 
The  slate bandit problem studied in this paper  can be used to model a reserve price optimization problem on a header bidding platform. The connection is as follows.
There are  \( M \) SSPs on the header bidding platform.
In every round \( t \) the publisher needs to choose a vector of reserve prices from the set \( \mathcal{B}  \). 
The revenue on  the header bidding platform when action \( b \in \mathcal{B}  \) chosen is given by \(  Y(b)   = f(Y_{1}(b_{1}), \dots,  Y_{M}(b_{M}))  = \max \lbrace Y_{1}(b_{1}), \dots,  Y_{M}(b_{M}) \rbrace \).  Note that Assumption~\ref{ass_independent_components_B} is  reasonable in this setting since (i) the pool of advertisers and their bidding strategies can differ across DSPs,  (ii) advertisers do not observe the bids (of their competitors) on other DSPs, and (iii) SSPs can be connected to different DSPs.

\section{Algorithms and Analysis} \label{sec:algorithm}
\subsection{The ETC-SLATE algorithm}
In this section we discuss our proposed algorithm. We refer to our algorithm as ETC-SLATE (Explore then Commit slate bandit algorithm).
The main idea that is used in our proposed algorithm relies on exploiting Assumption~\ref{ass_independent_components_B}. This is best illustrated using an example.

\begin{example}
Consider a simple instance of the slate bandit problem where there are \( M = 3 \) slots. Assume that \( \mathcal{B}_{1} = \lbrace x_{1}, x_{2} \rbrace \), \( \mathcal{B}_{2} = \lbrace y_{1}, y_{2} \rbrace \), \( \mathcal{B}_{3} = \lbrace z_{1}, z_{2} \rbrace \). Therefore, we have that \( \mathcal{B} = \lbrace  (x_{1}, y_{1}, z_{1}), (x_{1}, y_{1}, z_{2}), (x_{1}, y_{2}, z_{1}), (x_{1}, y_{2}, z_{2}), 
      (x_{2}, y_{1}, z_{1}), (x_{2}, y_{1}, z_{2}), (x_{2}, y_{2}, z_{1}), (x_{2}, y_{2}, z_{2})   \rbrace. \)

Suppose that, for every \( b \in  \mathcal{B} \),  we want to have \( N \) i.i.d. (independent and identically distributed) samples from \(  Y(b)   = f(Y_{1}(b_{1}), \dots,  Y_{3}(b_{3})) \) where \( b_{i} \in \mathcal{B}_{i}\). The straightforward way to do this is to collect \( N \) i.i.d. samples from \(  Y(b) \) by selecting every action \( b \in  \mathcal{B} \) exactly \( N \) times. Thus you would need \( N \cdot | \mathcal{B} | \) samples in total. 

A more efficient approach is simply to sample action \( (x_{1}, y_{1}, z_{1}) \) and action \( (x_{2}, y_{2}, z_{2}) \) exactly \( N \) times and save the values of \( Y_{1}(x_{1}) \), \( Y_{1}(x_{2}) \), \( Y_{2}(y_{1}) \), \( Y_{2}(y_{2}) \), \( Y_{3}(z_{1}) \), \( Y_{3}(z_{2}) \).  By Assumption~\ref{ass_independent_components_B}, we can use these samples to obtain \( N \) i.i.d. samples  from \(  Y(b)  \) for all  \( b \in  \mathcal{B} \). To get an i.i.d. sample from \(  Y((x_{2}, y_{1}, z_{2}))   = f(Y_{1}(x_{2}), Y_{2}(y_{1}),  Y_{3}(z_{2})) \), we simply use a sample from \( Y_{1}(x_{2}) \), \( Y_{2}(y_{1}) \) and  \( Y_{3}(z_{2}) \). Note that this approach only requires \( N \cdot 2 \) samples in total and this is less than the \( N \cdot | \mathcal{B} | \) samples of the previous approach. Note in particular that this approach allows us to obtain samples for actions  \( b \in  \mathcal{B} \) that have not been selected. In our example above, action \( (x_{2}, y_{1}, z_{2}) \) was not selected. However, by selecting action \( (x_{1}, y_{1}, z_{1}) \) and action \( (x_{2}, y_{2}, z_{2}) \) we do obtain the necessary information that allows us to construct an artificial i.i.d. sample from \(  Y((x_{2}, y_{1}, z_{2}))  \). \qed
\end{example}

The pseudo-code for ETC-SLATE is given by Algorithm~\ref{alg:FTC-bandit}. 
The main idea is to divide the horizon \( T \) into two phases. 
The first phase (the exploration phase) has length \( N = \hat{N}K  \) and the second phase (the commit phase) has length \( T - N \). 
In the first phase, the algorithm determines the best action \( \hat{b} \) in action 
set \( \mathcal{B}  \). In the second phase,  the algorithm commits to using action \( \hat{b} \) in each round.

In the first phase, the algorithm takes a subset \( \mathcal{B}^{F}  =  \lbrace \cup_{l  = 1}^K (l, \dots, l) | (l, \dots, l) \in \mathcal{B} \rbrace \) of actions from the set \( \mathcal{B}  \) and selects each action in this subset
\( \hat{N} \) times. Each time that action \( b \in  \mathcal{B}^{F} \) is selected, the rewards of the slots are observed (Line 6) and stored 
for later use (Line 7).
In Lines 11-16, the stored rewards for the slots are used in order to generate  \( \hat{N} \) i.i.d.  samples of the random variable
\(Y(b)   \) which are given by \( \hat{Y}^{1}(b), \dots, \hat{Y}^{\hat{N}}(b)  \). 
In Line 17,   the empirical mean of 
the \( \hat{N} \) values \( \hat{Y}^{1}(b), \dots, \hat{Y}^{\hat{N}}(b)  \) is determined for each action \( b \in \mathcal{B}  \).
The  action \( \hat{b} \) is then chosen as the action \( b \in \mathcal{B}  \) with the highest empirical mean.
The value of  \( \hat{N} \) is determined by the following  parameters: the horizon \(T \), \( \kappa \), \( \gamma \), and action set  \(\mathcal{B} \).
In Section~\ref{sec:regret_1} and \ref{sec:regret_2} we will show
that this choice for  \( \hat{N} \) leads to sub-linear regret for suitably chosen values of \( \kappa \) and \( \gamma \).

\begin{algorithm}[!ht!]
\begin{small}
\caption{ETC-SLATE  \label{alg:FTC-bandit}}
\begin{algorithmic}[1]
\REQUIRE  horizon \(T \), \( \kappa \), \( \gamma \), action sets  \(\mathcal{B} \).  \\
\STATE{Set \( \hat{N} = \bigg \lceil \frac{2}{{ \kappa}^{2}} \cdot  (\log{( |\mathcal{B}|)} - \log{(\gamma)} ) \bigg \rceil \). Set \(  t = 1\).}
%   \STATE{Set \(  t = 1\).} \\
 \STATE{Set \(\mathcal{V}_{i,j} = \varnothing \) \(  \forall \) \( i \in \lbrace 1, \dots, M \rbrace \) and \(  j \in \mathcal{B}_{i}  \). }

 \textbf{Explore Phase.}\\
  \FOR{ \( l \in \lbrace 1, \dots, K  \rbrace \) }
    \FOR{ \( n \in \lbrace 1, \dots, \hat{N} \rbrace \) }
 \STATE{Select action \( (l, \dots, l) \in \mathcal{B}   \).}
  \STATE{Observe rewards \( z_{l,i,n} \sim Y_{i}(l) \) for \( i = 1, \dots, M \). }
 \STATE{Set \(\mathcal{V}_{i,l} = \mathcal{V}_{i,l} \cup \lbrace  z_{l,i,n} \rbrace  \) for \( i = 1, \dots, M \). }
   \STATE{Set \(  t = t + 1\).} \\
  \ENDFOR \\
  \ENDFOR \\

  \FOR{ \( b = (l^{1}, \dots, l^{M})  \in \mathcal{B}  \) }
    \FOR{ \( n \in \lbrace 1, \dots, \hat{N} \rbrace \) }
 \STATE{Select    \(z_{l^{i},i,n} \in \mathcal{V}_{i,l^{i}} \) for \( i = 1, \dots, M \). } 
 \STATE{Set  \( \hat{Y}^{n}(b) = f(z_{l^{1},1,n}, \dots,  z_{l^{M},M,n}) \).}
  \ENDFOR \\
  \ENDFOR \\

 \textbf{Find best arm in \( \mathcal{B}\).}\\
 \STATE{Find \( \hat{b} \in \mathcal{B} \) such that \( \sum_{n = 1}^{\hat{N}} \hat{Y}^{n}(\hat{b}) \frac{1}{\hat{N}} \geq  \sum_{n = 1}^{\hat{N}} \hat{Y}^{n}(b) \frac{1}{\hat{N}} \) for all \( b \neq \hat{b}\). }  
  
 \textbf{Commit Phase.}\\
%   \STATE{Set \( \mathcal{C}^{-} = \mathcal{A} \cup \lbrace \hat{b}  \rbrace \).} \\
    %  \STATE{Set \(  t = \hat{N}K + 1\).} \\
  \FOR{ \( t \in  \lbrace \hat{N}K + 1, \dots, T  \rbrace \) }
 \STATE{Play action \(\hat{b} \). }
  \ENDFOR \\

\end{algorithmic}
\end{small}
\end{algorithm}

\subsection{Problem-dependent regret bounds} \label{sec:regret_1}

\begin{lemma}[\cite{Hoeffding}] \label{Hoeffding}
Let \( X_{1}, \dots, X_{n}\) be independent random variables such that  \( X_{i} \in [a, b]  \) for \( i = 1, \dots, n \). Let \(\bar{X} =  \frac{1}{n} \cdot \sum_{i = 1}^{n} X_{i}  \), and let \( \epsilon \geq 0  \).  Then, \( \mathbb{P} \left \lbrace   \bar{X}  - \mathbb{E} \left \lbrace   \bar{X} \right \rbrace    \geq \epsilon \right \rbrace \leq \exp \left \lbrace \frac{-2 { \epsilon}^{2} { n}^{2}}  {  n{(b - a )}^{2}} \right \rbrace \). 
\end{lemma}

\begin{proposition}[ ] \label{prop:prop2}
Let \( Y^{1}(b), \dots, Y^{n}(b) \) be \(n \) i.i.d draws  from \(Y(b)   \) for an action \( b \in \mathcal{B} \).  Assume that \(Y^{j}(b) \) and \(Y^{k}(l)\) are independent if \( b \neq l \) and \( j \neq k \). Let \( \bar{\mu}(b^{*}) = \sum_{i = 1}^{n}  Y^{i}(b^{*}) \cdot \frac{1}{n} \) and \( \bar{\mu}(b) = \sum_{i = 1}^{n}  Y^{i}(b) \cdot \frac{1}{n} \) for \( b \neq  b^{*} \). 
Let \( \hat{b} = \argmax_{b \in \mathcal{B}} \bar{\mu}(b)  \),  where ties are broken  arbitrarily if there are multiple candidates for  \( \hat{b} \). 
Then, \(   \mathbb{P}  \left \lbrace  \hat{b} \neq  b^{*}  \right \rbrace   \leq \bar{K} \exp \left \lbrace  -\frac{1}{2}n {(\Delta_{\min} )}^{2} \right \rbrace \). 

\begin{proof}
Define \( \mathcal{B}^{-} =  \lbrace b \in \mathcal{B} | \bar{\mu}(b) \geq \bar{\mu}(b^{*}), b \neq b^{*} \rbrace \).  Then we have that, 
\(  \mathbb{P}  \left \lbrace  \hat{b} \neq  b^{*}  \right \rbrace =  \mathbb{P}  \left \lbrace   \hat{b} \in \mathcal{B}^{-}   \right \rbrace 
   \stackrel{(a)}{\leq} \sum_{b \in \mathcal{B}^{-}}  \mathbb{P} \left \lbrace \hat{b} =  b  \right \rbrace 
 \stackrel{(b)}{\leq} \sum_{b \in \mathcal{B}^{-}}  \mathbb{P} \left \lbrace \bar{\mu}(b^{*}) \leq \bar{\mu}(b) \right \rbrace  
  \stackrel{(c)}{\leq} \sum_{b \in \mathcal{B}^{-}}  \exp \left \lbrace  -\frac{n}{2} {(\Delta(b))}^{2} \right \rbrace 
   \stackrel{(d)}{\leq} \bar{K} \exp \left \lbrace   - \frac{n}{2}  {(\Delta_{\min} )}^{2}\right \rbrace.\)

Inequality (a) follows from applying a union bound over the set \( \mathcal{B}^{-} \). Inequality (b) follows from the fact that  \( \mathbb{I} \left \lbrace \hat{b} =  b \right \rbrace = 1  \Rightarrow \mathbb{I} \left \lbrace \bar{\mu}(b^{*}) \leq \bar{\mu}(b) \right \rbrace = 1 \). 
Inequality (c) follows from applying Lemma~\ref{Hoeffding} to the differences \( Y^{i}(b^{*})  - Y^{i}(b)\) for \( i = 1, \dots, n \).
Inequality (d) follows from the fact that \( | \mathcal{B}^{-} | \leq \bar{K} = |\mathcal{B}| \) and \( {(\Delta_{\min} )}^{2} \leq {(\Delta(b))}^{2} \) for \( b \in \mathcal{B} \). 
\end{proof}
\end{proposition}

Recall that \( \hat{b} \) denotes the action in  \( \mathcal{B} \) that is identified as \( b^{*} \) by Algorithm~\ref{alg:FTC-bandit}. The following proposition bounds the probability that \( b^{*} \) is incorrectly identified.  
\begin{proposition}   \label{prop:prop3}
  Let \( m > 0 \). Let \( \hat{b} \) denote the action in  \( \mathcal{B} \) that is identified as \( b^{*} \) by Algorithm~\ref{alg:FTC-bandit}.  If Algorithm~\ref{alg:FTC-bandit} is run with the  inputs: \( T \),  \(\kappa =  \Delta_{\min} \), \( \gamma = \frac{1}{T^{m}} \), and action set  \( \mathcal{B} \),
then  \(  \mathbb{P}  \left \lbrace  \hat{b} \neq  b^{*}  \right \rbrace  \leq  \gamma \). 

\begin{proof}
 From the description of Algorithm~\ref{alg:FTC-bandit}, it follows that \( \hat{N} = 2 {(\Delta_{\min} )}^{-2} \cdot  (\log{(\bar{K})} - \log{(\gamma)} ) \).  Given this choice of \( \hat{N} \), we are able to generate \( \hat{N} \) i.i.d draws from \(Y(b)   \) for each \( b \in \mathcal{B} \). Let \( \hat{b} \) denote the action that has the highest empirical mean based on the \( \hat{N} \) samples and recall that \(b^{*} \) is the action with the highest expected return. By Proposition~\ref{prop:prop2} it follows that 
 \(
\mathbb{P}  \left \lbrace  \hat{b} \neq  b^{*}  \right \rbrace \leq \bar{K} \exp \left \lbrace  - \frac{\hat{N}}{2} {(\Delta_{\min} )}^{2} \right \rbrace  = \frac{1}{T^{m}}= \gamma \). 
\end{proof}
\end{proposition}

We can now state the main result of this subsection (Proposition~\ref{Thm: theorem1}). 

\begin{proposition} \label{Thm: theorem1}
Let \( m >0\). If Algorithm~\ref{alg:FTC-bandit} is run with inputs: \( T \),  \( \kappa=   \Delta_{\min} \), \( \gamma = \frac{1}{T^{m}} \), and action set \( \mathcal{B} \),
then \(  \mathcal{R}_{T} \leq  \frac{2K}{{(\Delta_{\min} )}^{2}}  \cdot  (\log{(\bar{K})} +  m\log{(T)} ) + T^{1-m}\) with  \( \bar{K} = | \mathcal{B} |  \).

\begin{proof}
Note that the regret \( \mathcal{R}_{T} \) can be decomposed as \(
 \mathcal{R}_{T}  =  \mathcal{R}_{N}  +  \mathcal{R}_{T-N}\).
Here \( \mathcal{R}_{N}  \) denotes the regret over the first \( N \) rounds and \( \mathcal{R}_{T-N}  \) denotes the regret over the last \( T-N \) rounds.
In order to bound \( \mathcal{R}_{T} \) it suffices to bound each term. 

\noindent \underline{Bounding \( \mathcal{R}_{N}  \)} \\
Note that \( \mathcal{R}_{N}  \)  is trivially bounded by \(  N \cdot 1  \) since by Assumption~\ref{ass_bounded_rewards} the regret for any period is at most \( 1 \). 
 From the description of Algorithm~\ref{alg:FTC-bandit}, it follows that \( \hat{N} = \frac{2}{{(\Delta_{\min} )}^{2}} \cdot  (\log{(\bar{K})} - \log{(\gamma)} ) \).   Given  \( \hat{N} \), it follows that Phase I has length \( N = K \cdot \hat{N}\). By substituting the quantities for \( \gamma  \) and \( \kappa\), we conclude that \(
 \mathcal{R}_{N}  \leq \frac{2K}{{(\Delta_{\min} )}^{2}}  \cdot  (\log{(\bar{K})} +  m\log{(T)} )\).

\noindent  \underline{Bounding \( \mathcal{R}_{T-N}  \)} \\
 We decompose \( \mathcal{R}_{T-N}  \) according to two cases:
 \begin{itemize}
\item [(i)]  \( \hat{b} \neq  b^{*} \). If case (i) occurs,  then \( \mathcal{R}_{T-N}  \)  is trivially bounded by \(  (T-N) \cdot 1  \). Therefore we conclude that in case (i) \( \mathcal{R}_{T-N} \leq T -N \).
\item [(ii)]  \( \hat{b} =  b^{*} \). If case (ii) occurs,  then \( \mathcal{R}_{T-N}  = 0 \). This follows from the fact that, \( \Delta(b^{*})   = 0 \). 
 \end{itemize}
 By combining the results for the two cases above and noting that by Proposition~\ref{prop:prop3} we have \(  \mathbb{P}  \left \lbrace  \hat{b} \neq  b^{*}  \right \rbrace  \leq  \gamma \), we obtain \(
 \mathcal{R}_{T-N}  \leq (T-N) \cdot  \mathbb{P}  \left \lbrace  \hat{b} \neq  b^{*}  \right \rbrace  + \mathbb{P}  \left \lbrace  \hat{b} =  b^{*}  \right \rbrace \cdot 0 \\
  \leq (T-N) \cdot  \frac{1}{T^{m}} \leq T \cdot  \frac{1}{T^{m}} = T^{1-m}\).
\end{proof}
\end{proposition}

\begin{corollary} \label{Corr:FTC_corr1}
 Let \( \bar{K} = | \mathcal{B} | \leq T  \) and \( m = 1 \). Suppose that  Algorithm~\ref{alg:FTC-bandit} is run with inputs: \( T \),  \(  \kappa = \Delta_{\min} \), \( \gamma = \frac{1}{T^{m}} \),   and action set \( \mathcal{B} \). Then,  \( \mathcal{R}_{T} \leq \frac{2K}{{(\Delta_{\min} )}^{2}}  \cdot  ( 2\log{(T)} ) + 1\).
\end{corollary}

If \( \Delta_{\min} \)  is not precisely known, we can still run Algorithm~\ref{alg:FTC-bandit}  using a lower bound for \(\Delta_{\min}\) if this is available. In Proposition~\ref{Thm: theorem1}, the dependence of regret on \(T \) would then still be logarithmic in \(T \) but with a different problem-dependent constant.

\subsection{Problem-independent regret bounds} \label{sec:regret_2}
The results of the previous section show that expected regret of order \( O( \log{T}) \) is possible if the gaps are known. However, as \( \Delta_{\min} \rightarrow 0 \), the regret bounds in Proposition~\ref{Thm: theorem1} and Corollary~\ref{Corr:FTC_corr1} becomes vacuous. Therefore, it is useful to study whether sub-linear regret is possible when the gaps are unknown. In this section we prove  problem-independent regret bounds and show that sub-linear regret is still achievable.

\begin{proposition} \label{Thm: theorem_gap_independent}
 Let \( m > 0 \). If Algorithm~\ref{alg:FTC-bandit} is run with inputs: \( T \),  \( \kappa = T^{-1/3} \sqrt{K} \sqrt{\log{(T)}} \), \( \gamma = \frac{1}{T^{m}} \), and action set \( \mathcal{B} \),
then \( \mathcal{R}_{T} \leq  \frac{2T^{2/3}}{\log{(T)}} \cdot  (\log{(\bar{K})} +  m\log{(T)} ) + T^{1-m} 
 +  T^{2/3} \sqrt{K} \sqrt{\log{(T)}}\)
  with \( \bar{K} = | \mathcal{B} |  \).

\begin{proof}
\ The proof uses similar arguments as in Proposition~\ref{prop:prop3} and \ref{Thm: theorem1}. Define the set \(  \mathcal{B}^{H} = \lbrace b \in \mathcal{B}| \Delta{(b)} \geq   \kappa  \rbrace \). Let \( \hat{b} \) denote the action in  \( \mathcal{B} \) that is identified as \( b^{*} \) by Algorithm~\ref{alg:FTC-bandit}. Using similar arguments as in the proof of Proposition~\ref{prop:prop3}, we conclude that \(  \mathbb{P}  \left \lbrace  \hat{b} \in  \mathcal{B}^{H} \right \rbrace  \leq  \gamma \). 
We again decompose the regret \( \mathcal{R}_{T} \) as \( \mathcal{R}_{T}  =  \mathcal{R}_{N}  +  \mathcal{R}_{T-N}\).

\noindent \underline{Bounding \( \mathcal{R}_{N}  \)} \\
Note that \( \mathcal{R}_{N}  \)  is trivially bounded by \(  N \cdot 1  \) since by Assumption~\ref{ass_bounded_rewards} the regret for any period is at most \( 1 \). 
From the description of Algorithm~\ref{alg:FTC-bandit}, it follows that \( \hat{N} = 2{\kappa}^{-2} \cdot  (\log{(\bar{K})} - \log{(\gamma)} ) \).    Given  \( \hat{N} \), it follows that Phase I has length \( N = K \cdot \hat{N}\). By substituting the quantities for \( \gamma  \) and \( \kappa\), we conclude that 
\( \mathcal{R}_{N}  \leq \frac{2T^{2/3}}{\log{(T)}} \cdot  (\log{(\bar{K})} +  m\log{(T)} )\).

\noindent  \underline{Bounding \( \mathcal{R}_{T-N}  \)} \\
 We decompose \( \mathcal{R}_{T-N}  \) according to two cases:
 \begin{itemize}
\item [(i)]  \( \hat{b} \in  \mathcal{B}^{H} \). If case (i) occurs,  then \( \mathcal{R}_{T-N}  \)  is trivially bounded by \(  (T-N) \cdot 1  \).  
% This follows from the observation that \( c^{*} \) has optimality gap of zero, so that the regret could be \( 1 \) for each period in the worst case. 
Therefore, we conclude that in case (i) \( \mathcal{R}_{T-N} \leq T -N \).
\item [(ii)]  \( \hat{b} \notin  \mathcal{B}^{H} \).    If   \( \hat{b} \notin  \mathcal{B}^{H} \), then from the definition of \(\mathcal{B}^{H} \), it follows that \(  \Delta(\hat{b})  \leq \kappa \). 
\end{itemize}
 By combining the results for the two cases above and noting that  \(  \mathbb{P}  \left \lbrace  \hat{b} \in  \mathcal{B}^{H} \right \rbrace  \leq  \gamma \), we obtain \\
 \(
 \mathcal{R}_{T-N}  \leq (T-N) \cdot  \mathbb{P}  \left \lbrace  \hat{b} \in  \mathcal{B}^{H}  \right \rbrace  + \mathbb{P}  \left \lbrace  \hat{b} \notin  \mathcal{B}^{H}  \right \rbrace \cdot (T-N) \kappa  \\
 \leq T \cdot  \frac{1}{T^{m}}  + T \kappa = T^{1-m} + T \kappa\). 
\end{proof}
\end{proposition}

\begin{corollary} \label{Corr:Regret_UCB_problem_independent}
 Let \( \bar{K} = | \mathcal{B} | \leq T  \) and \( m = 1 \). Suppose that  Algorithm~\ref{alg:FTC-bandit} is run with inputs: \( T \),  \( \kappa= T^{-1/3} \sqrt{K} \sqrt{\log{(T)}} \sqrt{(1 + m)}  \), \( \gamma = \frac{1}{T^{m}} \),  and action set \( \mathcal{B} \). Then, 
  \(   \mathcal{R}_{T} \leq    T^{2/3} \cdot (2 + \sqrt{2K\log{(T)}}) + 1\). 
\end{corollary}

It is useful to compare the obtained bounds with previously known results. If we consider every slate as a separate action in a standard multi-armed bandit algorithm such as UCB1, then regret of order \( O(\sqrt{T\log{(T)}} \sqrt{K^{M}}) \) is possible \cite{MAL-024,MAL-038}. If we compare this with Corollary~\ref{Corr:Regret_UCB_problem_independent}, then we have a worse dependence on \( T \) (we have \( T^{2/3} \sqrt{\log{(T)}} \) instead of \(\sqrt{T\log{(T)}} \)) but a better dependence on \( K \). It is an open problem whether the dependence on \( T \) can  be improved further.

\section{Experiments}
\label{sec:experiments}
\FloatBarrier
In this section we conduct experiments in order to test the performance of our proposed algorithm. We conduct experiments using both simulated data and real-world data. 

\subsection{Experiments using simulated data}
\label{sec:experiments_SIM}
The main purposes of the experiments with simulated data are to verify the theoretical results that were derived, and to investigate the effects of ignoring the non-separability of the slate-level reward function on the regret. 

\subsubsection{Experimental settings}
In the experiments we set  \(  M = 5 \) and   \( \mathcal{B}_{i}  = \lbrace 1, \dots, 10 \rbrace  \) for \( i = 1, \dots, M \). We consider three choices for the slate-level reward function. These choices are: 
\begin{itemize}
    \item \( f_{1} =  \frac{1}{4} \max \lbrace Y_{1}(b_{1}), Y_{2}(b_{2})  \rbrace  + \frac{1}{4} \max \lbrace Y_{2}(b_{2}), Y_{3}(b_{3})  \rbrace + \frac{1}{4} \max \lbrace Y_{3}(b_{3}) , Y_{4}(b_{4})  \rbrace  + \frac{1}{4} \max \lbrace Y_{4}(b_{4}), Y_{5}(b_{5})  \rbrace \)
    \item   \( f_{2} =  \frac{1}{4}\max \lbrace Y_{1}(b_{1}), Y_{2}(b_{2})  \rbrace  + \frac{1}{4} Y_{3}(b_{3})   + \frac{1}{4}   Y_{4}(b_{4})    + \frac{1}{4} \max \lbrace Y_{4}(b_{4}), Y_{5}(b_{5})  \rbrace \), 
    \item  \( f_{3} =  \frac{1}{4}\max \lbrace Y_{1}(b_{1}), Y_{2}(b_{2})  \rbrace  + \frac{1}{4} \max \lbrace Y_{1}(b_{1}), Y_{3}(b_{3})  \rbrace + \frac{1}{4} \max \lbrace Y_{1}(b_{1}), Y_{4}(b_{4})  \rbrace  + \frac{1}{4} \max \lbrace Y_{1}(b_{1}), Y_{5}(b_{5})  \rbrace \). 
\end{itemize}

In our experiments the rewards for \( b \in  \mathcal{B}_{i}\) follow a uniform distribution on \(  [a-c, a + c] \) where \( a \) is chosen uniformly from \( [0.4, 0.6] \) independently for \( i = 1, \dots, M \) and for all \( b \in   \mathcal{B}_{i} \), and \( c \) is chosen uniformly from \( [0.1, 0.3] \) independently from \( a \). In total we have three experimental settings: Exp1, Exp2, Exp3. The abbreviation Exp1 means that \( f_{1} \) is used. The other abbreviations have a similar interpretation. \\
The main motivation for the choice of slate-level reward functions  and the reward distributions is that, the slate-level reward functions are non-separable, but since the reward distributions are uniform, the  optimal slate and the regret can still be calculated analytically. 

In the experiments, ETC-SLATE is tuned according to Corollary~\ref{Corr:Regret_UCB_problem_independent}. To the best of our knowledge, there are no existing algorithms for our slate bandit problem with non-separable rewards.
For this reason we used the following benchmark. We run a standard multi-armed bandit algorithm on the base actions at the slot-level (for each slot independently), and we then combine the base actions chosen by these independent bandits in order to form the action at the slate-level. This is a reasonable benchmark, in the sense that assuming a non-decreasing reward function at the slate-level, this should allow this benchmark to learn the optimal action over time. In the experiments we use the UCB1 \cite{Auer2002} and Thompson sampling (TS) \cite{pmlr-v31-agrawal13a_short} as the multi-armed bandit algorithms at the slot-level.

\subsubsection{Results}
\begin{figure} [!h]
\centerline{    \includegraphics[width=1.0\columnwidth]{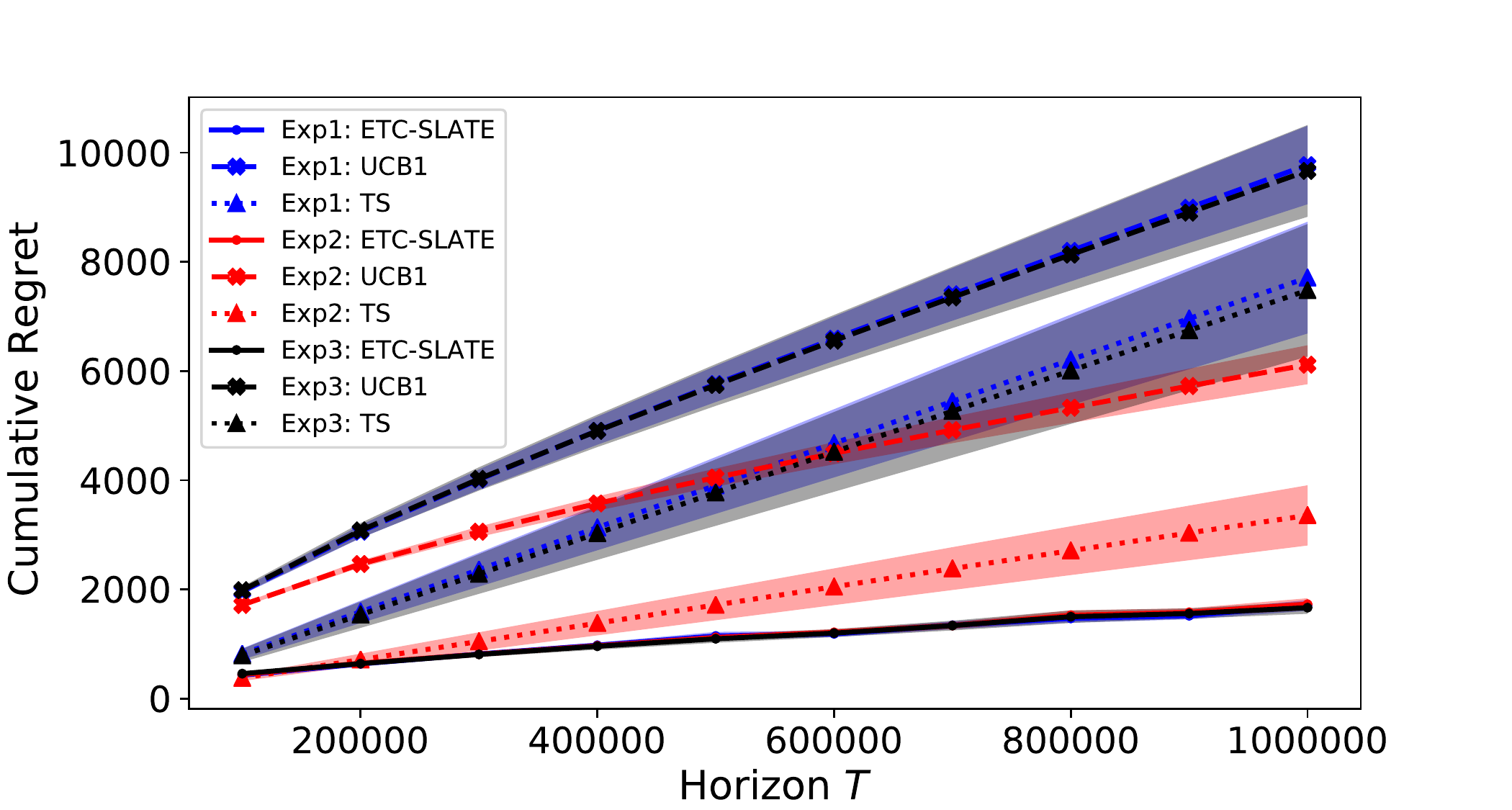}}
\caption{Performance of algorithms averaged over 200 runs. Lines indicate the mean and shaded region indicates 95\% confidence interval.} \label{fig:Cum_regret_SIM_A}
\end{figure}

In Figure~\ref{fig:Cum_regret_SIM_A}  the cumulative regret is shown for different experimental settings and different values for the problem horizon. Each point in the graph shows the cumulative regret over \( T \) rounds for a slate bandit problem of horizon \( T \) averaged over 200 simulations. The results indicate that ETC-SLATE clearly outperforms the  benchmarks. The regret of UCB1 is at least twice as high as ETC-SLATE.  TS tends to outperform UCB1, but the regret is still at least 40\% higher compared to   ETC-SLATE. Also, we note that ETC-SLATE performs similarly on all the test functions, but for UCB1 and TS the performance on \( f_{1} \) and \( f_{3} \) differs from \( f_{2} \). 
The results in Figures~\ref{fig:Cum_regret_SIM_A}  also confirm that the regret bound from Corollary~\ref{Corr:Regret_UCB_problem_independent} indeed holds. However, by comparing the regret curve  with the expression for the regret bound,  it appears that the bound is not  tight and this suggests that the bound could be improved further. 

\subsection{Experiments using real-world data}
\label{sec:experiments_RTB}
In this section we perform experiments on the reserve price optimization problem with header bidding. In this problem, there are  \( M \) SSPs on the header bidding platform. In every round \( t \), the publisher needs to choose a reserve price \(b_{i} \) from the set \( \mathcal{B}_{i}  \). 
The revenue on  the header bidding platform when action \( b \in \mathcal{B}  \) chosen is given by \(  Y(b)   = f(Y_{1}(b_{1}), \dots,  Y_{M}(b_{M}))  = \max \lbrace Y_{1}(b_{1}), \dots,  Y_{M}(b_{M}) \rbrace \).

\subsubsection{Dataset description}
In order to evaluate our method we use real-life data from ad auction markets from the publicly available iPinYou dataset \cite{zhang2014real}. It contains bidding information from the perspective of nine advertisers on a Demand Side Platform (DSP) during a week.  The dataset  contains information about the top bid and the second bid if the advertiser wins an auction.  We use the iPinYou dataset to construct synthetic data for the top bid and second bid in order to test our proposed approach.

% We now describe how we construct the joint distribution of the top two bids for each SSP. 
We use data from the advertisers to model the bids from an SSP. Fix an advertiser (say advertiser \( m \)) and fix an hour of the day (say hour \( h \)).  
For   advertiser \( m \) we take the  values of the second highest bid in hour \( h \) and we filter these values  by the ad exchange (there are two ad exchanges) on which the bids were placed.  Next, we sample (with replacement) 10000 values for each ad exchange to approximate the distribution of the second bid. After these steps we end up with 2 lists \( L_{m,h,1} \) and \( L_{m,h,2} \) of size 10000 for each ad exchange for advertiser \(m \) in hour \( h \). Define \( L^{max}_{m, h} \) as the maximum value of all values in \( L_{m,h,1} \) and \( L_{m,h,2} \). 
We use the following procedure to construct the bids for a horizon  of length \( T \).  For round \( t \in \lbrace 1, \dots, T\rbrace \)  we draw \( A_{t} \) uniformly at random from \( L_{m,h,1} \) and \( B_{t} \) uniformly at random from \( L_{m,h,2} \). The highest bid in round \( t \) is given by \( X_{t} =   \max \lbrace A_{t}, B_{t} \rbrace / L^{max}_{m, h} \) and the second-highest bid is given by \( Y_{t} = \min \lbrace A_{t}, B_{t} \rbrace / L^{max}_{m, h} \).  Denote the resulting joint distribution by \( D_{m,h} \). 

\subsubsection{Experimental settings}
In the experiments we assume that there are \(  M = 4 \) SSPs.
The action sets \( \mathcal{B}_{i} \) for SSP \(i \)  is given by \( K = 15 \) reserve prices  which are equally spaced in the interval \(    [0.1, 0.8] \). 
We consider three experimental settings and in each setting the distributions \( D_{m,h} \) are different. The different experimental settings are summarized as follows.
\begin{itemize}
    \item Setting Exp1. In Exp1 we use data from advertisers 1458, 3358, 3386 and  3427 on day 2 and from hour 18. 
    \item Setting Exp2. In Exp2 we use data from advertisers 1458, 3358, 3386 and  3427 on day 2 and from hour 15. 
    \item Setting Exp3. In Exp3 we use data from advertisers 1458, 2261, 2821 and  3427 on day 3 and from hour 18. 
\end{itemize}

In order to measure the performance of the methods we look at the per period reward, which  is defined as \(PPR(T) =  \sum_{t = 1}^{T} \hat{R}_{t}  / T \).  Here \( \hat{R}_{t} \) is the observed reward in round \( t \). 

In the experiments, ETC-SLATE is tuned according to Corollary~\ref{Corr:Regret_UCB_problem_independent} and we use the same benchmarks as in the previous experiments.

\subsubsection{Results}
\begin{figure} [!htb]
\centerline{    \includegraphics[width=0.75\columnwidth]{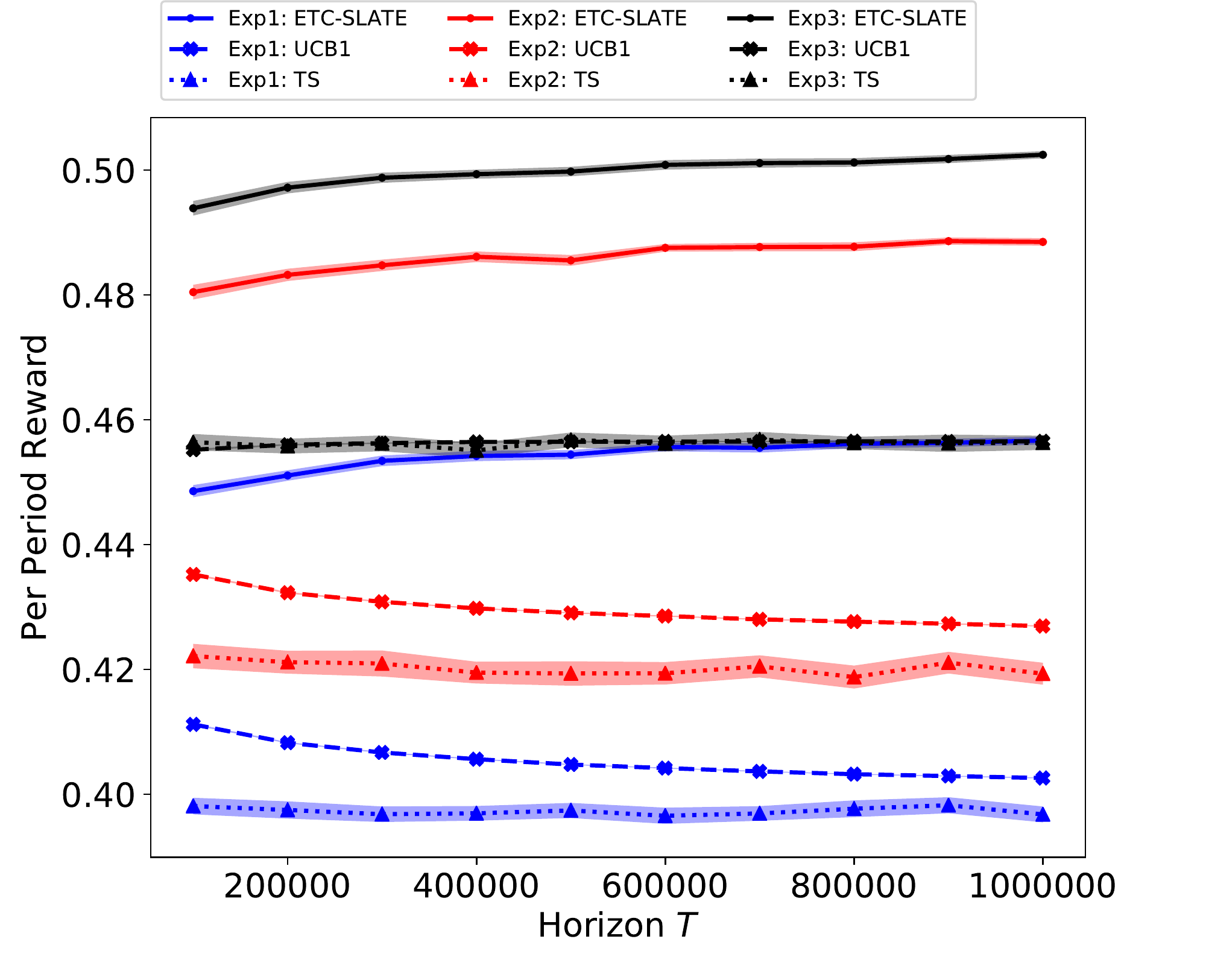}}
\caption{Performance of algorithms averaged over 200 runs. Lines indicate the mean and shaded region indicates 95\% confidence interval.} \label{fig:PPR_RTB}
\end{figure}

Figure~\ref{fig:PPR_RTB} shows the per period reward. From this figure we  observe that the difference in performance is  quite substantial as ETC-SLATE has a cumulative reward that is on average 10\% higher than  the benchmarks. Furthermore, the results indicate that the  difference in performance is not sensitive with respect to the underlying distributions at the slot-level.

\section{Conclusion}
\label{sec:conclusion}
In this paper we study slate bandits with non-separable reward functions at the slate-level.
Previous papers have only considered the case where the slate-level reward satisfies a monotonicity property.
The non-separability property implies that choosing the optimal base action   for each slot does not necessarily lead  to the highest expected reward at the slate-level. 
We provide a theoretical analysis and derive problem-dependent and problem-independent regret bounds.  
Furthermore, we provide   algorithms that  have sub-linear regret with respect to the time horizon. 

The work presented in this paper can be improved in a number of ways.
In our analysis we made the assumption that the slot-level rewards are independent from each other. However,
other papers (e.g. \cite{Dimakopoulou}) do not make such assumptions. 
It is not clear how to tackle the slate bandit problem in such a situation and future research can be directed
towards deriving sub-linear regret bounds for this case.

 \newpage

%%%%%%%%%%%%%%%%%%%%%%%%%%%%%%%%%%%%%%%%%%%%%%%%%%%%%%%%%%%%%%%%%%%%%%

%  \bibliographystyle{abbrv}
%  \bibliography{references_14april2020}{}

\end{document}